\documentclass[letterpaper]{article} 
\usepackage[]{aaai25}  
\usepackage{times}  
\usepackage{helvet}  
\usepackage{courier}  
\usepackage[hyphens]{url}  
\usepackage{enumitem}
\usepackage{amsmath}
\usepackage{xcolor}
\usepackage{booktabs}
\usepackage{graphicx} 
\usepackage{natbib}  
\usepackage{caption} 
\usepackage{algorithm}
\usepackage{algorithmic}

\usepackage{newfloat}
\usepackage{listings}
\DeclareCaptionStyle{ruled}{labelfont=normalfont,labelsep=colon,strut=off} 
\floatstyle{ruled}
\newfloat{listing}{tb}{lst}{}
\floatname{listing}{Listing}
\title{Language Ranker: A Metric for Quantifying LLM Performance Across \\ High and Low-Resource Languages}
\author{
    Zihao Li\textsuperscript{\rm 1}\thanks{\, Work done during Zihao Li's remote internship at NJIT.}, 
    Yucheng Shi\textsuperscript{\rm 2}, 
    Zirui Liu\textsuperscript{\rm 3}, 
    Fan Yang\textsuperscript{\rm 4}, 
    Ali Payani\textsuperscript{\rm 5}, 
    Ninghao Liu\textsuperscript{\rm 2}, 
    Mengnan Du\textsuperscript{\rm 1}
}
\affiliations{
    \textsuperscript{\rm 1}New Jersey Institute of Technology,
    \textsuperscript{\rm 2}University of Georgia,
    \textsuperscript{\rm 3}University of Minnesota at Twin Cities\\
    \textsuperscript{\rm 4}Wake Forest University,
    \textsuperscript{\rm 5}Cisco Research\\
    lizihao9885@gmail.com,\,\,yucheng.shi@uga.edu,\,\ zrliu@umn.edu,\,\, yangfan@wfu.edu,\,\, apayani@cisco.com,\,\,ninghao.liu@uga.edu,\,\,mengan.du@njit.edu
}

\usepackage{bibentry}

\begin{document}

\maketitle

\begin{abstract}
The development of Large Language Models (LLMs) relies on extensive text corpora, which are often unevenly distributed across languages. This imbalance results in LLMs performing significantly better on high-resource languages like English, German, and French, while their capabilities in low-resource languages remain inadequate. Currently, there is a lack of quantitative methods to evaluate the performance of LLMs in these low-resource languages. To address this gap, we propose the \emph{Language Ranker}, an intrinsic metric designed to benchmark and rank languages based on LLM performance using internal representations. By comparing the LLM's internal representation of various languages against a baseline derived from English, we can assess the model's multilingual capabilities in a robust and language-agnostic manner. Our analysis reveals that high-resource languages exhibit higher similarity scores with English, demonstrating superior performance, while low-resource languages show lower similarity scores, underscoring the effectiveness of our metric in assessing language-specific capabilities. 
Besides, the experiments show that there is a strong correlation between the LLM’s performance in different languages and the proportion of those languages in its pre-training corpus. 
These insights underscore the efficacy of the Language Ranker as a tool for evaluating LLM performance across different languages, particularly those with limited resources.

\end{abstract}

\section{Introduction}

Large Language Models (LLMs), such as GPT-4, Claude-3 and LlaMa-3, have demonstrated remarkable performance in various NLP tasks~\cite{achiam2023gpt,ouyang2022training,touvron2023llama,team2024gemma,jiang2023mistral,bai2023qwen}. However, this progress masks a critical issue: the stark disparity in LLM performance across languages, particularly disadvantaging low-resource languages common in developing countries. This disparity stems from the overwhelming bias towards high-resource languages, especially English, in training datasets~\cite{xie2024weakly}. For instance, approximately 92.65\% of GPT-3's training tokens are in English, leaving a mere 7.35\% for all other languages~\cite{gpt3_dataset_stats}. Similarly, English accounts for 89.70\% of LlaMa2's pre-training data~\cite{touvron2023llama}.

This imbalance leads to significant performance gaps between high-resource languages (e.g., English, German, French) and low-resource languages, potentially exacerbating global digital divides. LLMs often struggle with context-specific interpretations in low-resource languages, such as understanding culturally specific expressions or idioms~\cite{zhang2023don}. Recent studies confirm that pre-trained models perform poorly in languages with insufficient training data~\cite{lankford2024transformers}, highlighting the urgent need for robust methods to quantify and address these disparities. It is urgent to develop a tool to quantify linguistic biases in LLMs, so as to contribute to more equitable AI technologies, potentially improving access and performance for underserved language communities worldwide.

To tackle this challenge, we introduce \emph{Language Ranker}, a novel metric designed to systematically evaluate LLM performance across diverse languages, with a particular focus on low-resource languages. Our method leverages internal representations of LLMs, establishing English corpus representations as a baseline and measuring the similarity between this baseline and representations from other languages. This similarity score serves as a quantitative measure of language-specific performance. We validate Language Ranker by applying it to five state-of-the-art LLMs: {LlaMa2}~\cite{touvron2023llama}, {LlaMa3}~\cite{llama3github}, {Qwen}~\cite{bai2023qwen}, {Mistral-v0.1}~\cite{jiang2023mistral}, and {Gemma}~\cite{team2024gemma}. Additionally, we analyze the relationship between Language Ranker scores, the proportion of languages in training datasets, and performance on established benchmarks.
Our comprehensive experiments yield the following key findings:

\begin{itemize}[leftmargin=*]\setlength\itemsep{-0.3em}
\item Experimental results indicate that high-resource languages consistently show higher similarity scores with English, while low-resource languages exhibit lower scores, validating Language Ranker's effectiveness in quantifying language-specific performance disparities.
\item We uncover a strong correlation between LLM performance and the proportion of languages in pre-training corpora, providing crucial insights into the impact of training data distribution on multilingual capabilities.
\item The analysis of embedding space distributions reveals that high-resource languages are more evenly distributed, while low-resource languages cluster narrowly, further supporting the reliability of the proposed metric.
\end{itemize}

\section{The Proposed Method}

In this section, we will give an introduction to our analysis method.
First, we will introduce the dataset that we used. Then, we will introduce how to obtain the similarity between English and other languages, as well as how to compare different LLMs' performances. 
\footnote{Our code link: \url{https://github.com/lizh9885/Language-Ranker/}}

\subsection{Probing Datasets}

We use OPUS-100~\cite{zhang-etal-2020-improving} as our evaluation datasets. OPUS-100 is an English-centric multilingual corpus that covers 100 languages. Each sample consists of text in a non-English language as the original data, with its English translation serving as the target data. For example, \{"German": "Ich wollte dir erst noch etwas zeigen.","English": "I wanted to show you something first."\}. After filtering, there are 94 subsets containing English, including high-resource languages such as German, French, and Chinese, as well as low-resource languages such as Oriya, Kannada, and Kazakh. Each subset contains 2000 samples.

\subsection{Similarity Measurement}
We employ cosine similarity to measure the LLMs' performance gap between the target language and English. Specifically, given two sentences $X = \{x_i\}_{i=1}^{n}$ and $Y = \{y_i\}_{i=1}^{m}$ representing the text in English and the text in the target language. Since the current LLMs are all autoregressive models, we use the representation obtained after LLM mapping of the last token $x_n$ and $y_m$ as the representation of the text and calculate the similarity between them.
As we know, LLM consists of several layers of Transformer blocks~\cite{vaswani2017attention}. Therefore, after each layer of mapping by the transformer block, we can get a representation vector $x_{n}^{l}$ and $y_{m}^{l}$, $l=1...H$, where $H$ represents the number of the layer of LLMs. 
According to \cite{li2024inference}, the intermediate representation can be briefly summarized by the following equations:
\begin{equation}
\small
    x^{l+1} = \text{MLP}(x^{l} + \text{MHA}(x^{l})) \quad l =1...H,
\end{equation}
where MHA means multi-head attention or multi-group attention, and MLP means standard multilayer perceptron layer. Next, we take $x_{n}^{l}$ and $y_{m}^{l}$ to calculate the similarity. To implement a more robust similarity measure, we use the average similarity obtained by several intermediate layers as the final similarity.  
This process can be described as follows:
\begin{equation}
\begin{aligned}
\footnotesize
Sim = \frac{1}{|l_{sub}|} \sum_{i=1}^{|l_{sub}|} Sim_{i}, \, \text{where} \,
Sim_{i} = \frac{x_{n}^{i} y_{m}^{i}}{||x_{n}^{i}||||y_{m}^{i}||},
\end{aligned}
\end{equation}
where $l_{sub}$ = $\{ 5, 10, 15, 20, 25\}$ is the subset of the layers we selected.
Finally, we use $Sim$ 
to evaluate the performance gap between English and Non-English corpus.
\subsection{Rank Correlation Measurement}

When we get the similarity between each non-English representation and the English representation, we sort them according to the similarity to get a sorted ranking list of all languages. To measure the similarity of the sorted ranking lists of two LLMs, we use the longest common partial order sublist to measure.
It can be defined as follows:
For two sorted lists $A$ and $B$, find a sublist $C$ that is a subset of $A$ and $B$ such that for any number of index $i_1 \leq i_2\leq...\leq i_n$, Index($C_{i_1}$)$\leq$Index($C_{i_2}$)$\leq$...$\leq$Index($C_{i_n}$) is true for both A and B, and the longest sublist C that makes it true is called the longest common partial order sublist of A and B.
We use the ratio of the length of the longest common partial order sublist of two LLMs to the total length of the ranking list as a metric to measure the correlation.

\begin{figure*}
    \centering
    \vspace{-10pt}
    \includegraphics[width=1.0\linewidth]{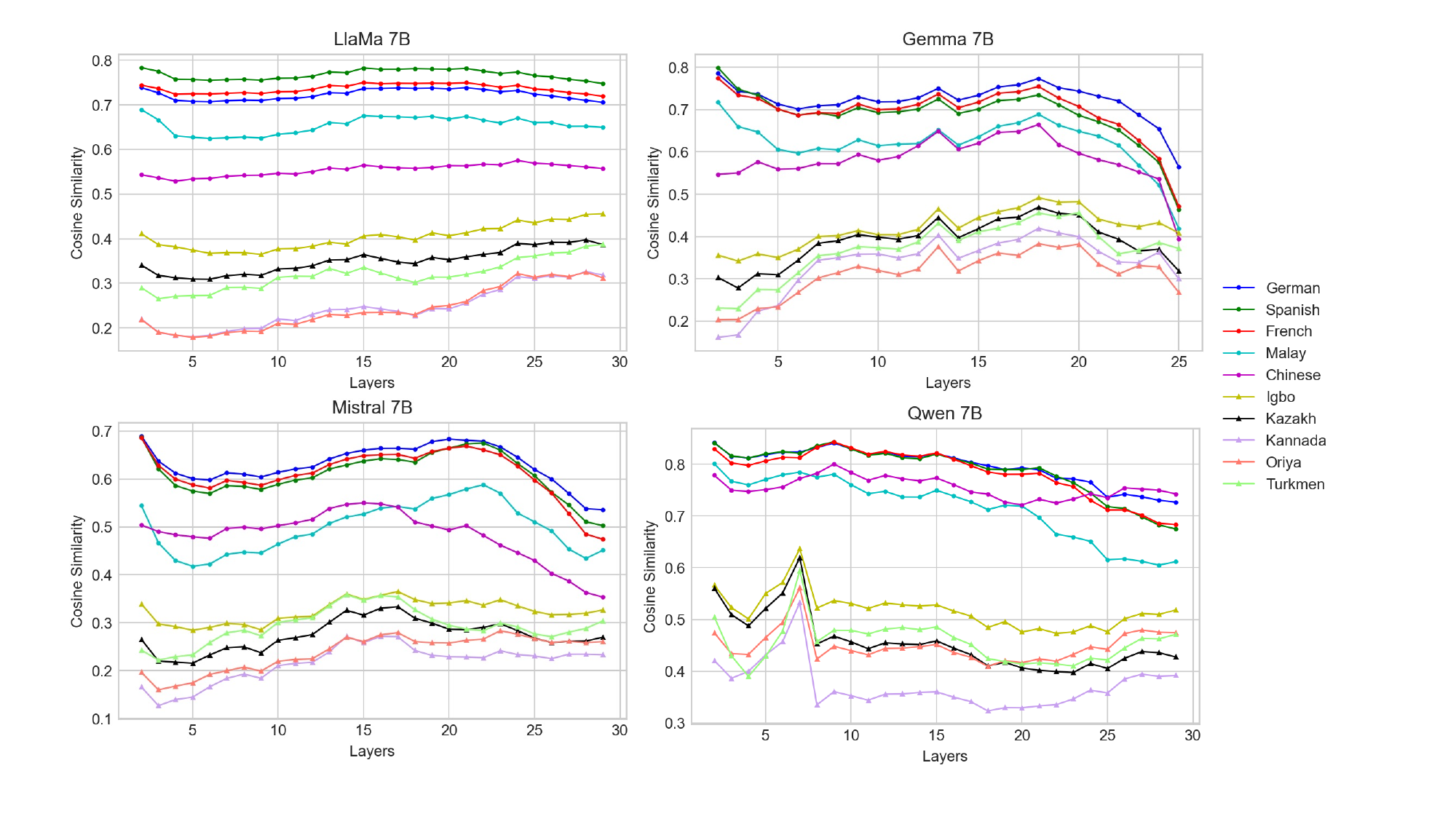}
    \caption{Performance of different LLMs for ten kinds of language. High-resource languages: German, Spanish, French, Indonesian and Chinese; and low-resource languages: Igbo, Kazakh, Kannada, Oriya and Turkmen.}
    \label{fig:main_label}
\end{figure*}

\section{Experiments}
In our experiments, we utilize five prominent open-source large models: {LlaMa2}~\cite{touvron2023llama}, {LlaMa3}~\cite{llama3github}, {Qwen}~\cite{bai2023qwen}, {Mistral-v0.1}~\cite{jiang2023mistral}, and {Gemma}~\cite{team2024gemma}.

We conduct experiments to answer the following research questions in the following four subsections respectively: RQ1: Can the Language Ranker effectively quantify the performance of LLMs across multiple languages? 
RQ2: How consistent are the performance rankings of different LLMs when evaluated across a diverse set of languages? 
RQ3: Is the proposed cosine similarity metric correlated with the proportion of a language in the LLMs' pre-training corpus? 
RQ4: Is the proposed cosine similarity metric correlated with performance on other benchmark tasks for quantifying the multilingual capabilities of LLMs?

\subsection{Can Language Ranker Quantify LLM Performance Across Languages? (RQ1)}
\label{3.1}
To visualize the performance of different LLMs in these languages, we selected 10 representative languages to display their inference results. They consist of five high-resource languages, including German, Spanish, French, Indonesian, and Chinese, and five low-resource languages, including Igbo, Kazakh, Kannada, Oriya, and Turkmen. Figure~\ref{fig:main_label} shows detailed results, where the X-axis represents different layers of LLMs, while the Y-axis represents the similarity between the target language and English for each layer.
From Figure~\ref{fig:main_label}, we can observe that high-resource languages have representations more similar to English, whereas low-resource languages show less similarity. 
Specifically, German, Spanish, French, and Malay generally maintain cosine similarity scores above 0.6, with Spanish and French often showing the highest scores, indicating that these languages are better represented in the models' embeddings. In contrast, low-resource languages, such as Igbo, Kazakh, Kannada, Oriya, and Turkmen, display significantly lower cosine similarity scores, often below 0.4. These results show the disparities in performance across languages and highlights the utility of the Language Ranker in quantifying these differences robustly.

\begin{figure}[t]  
  \centering  
  \includegraphics[width=0.45\textwidth]{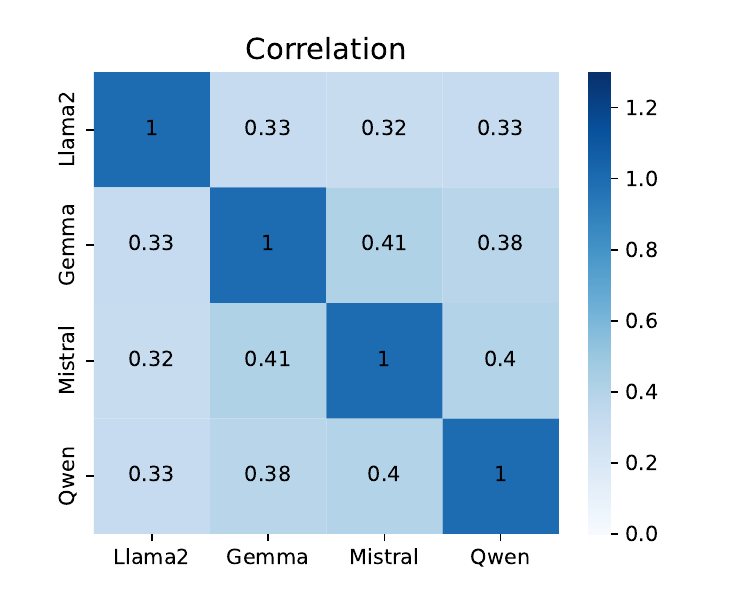}  
  \caption{Rank correlation between different LLMs. This is calculated using metric introduced in Section 2.3. It shows high correlations across LLMs.}  
  \label{fig:heat}  
\end{figure}  

\begin{figure}[htbp]
    \centering
    \vspace{-10pt}
    \includegraphics[width=1.0\linewidth]{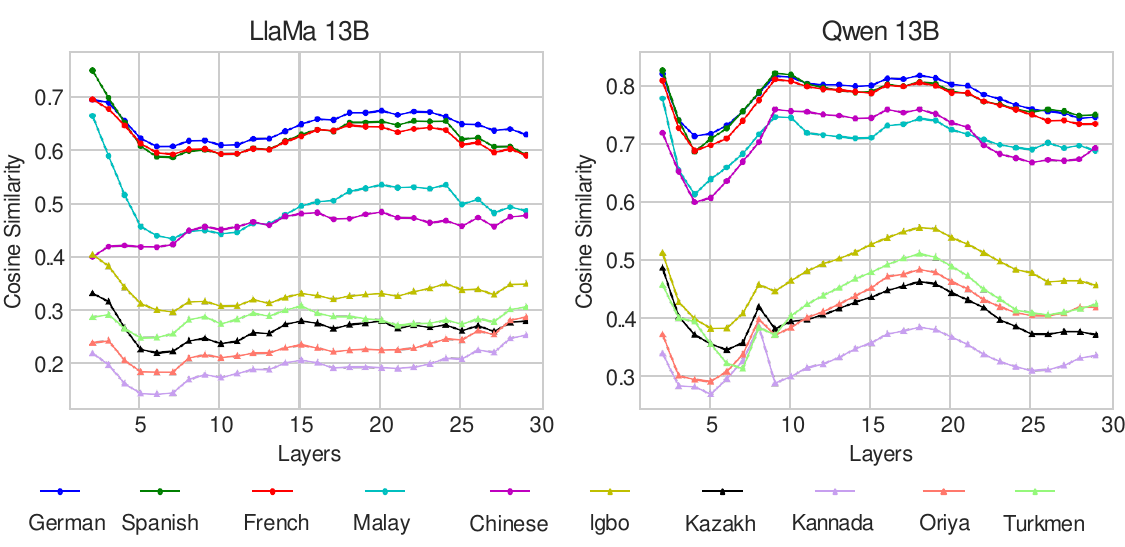}
    \caption{Cosine similarity scores across model layers for LlaMa2 13B and Qwen 13B. The graph shows results for the same 10 languages as in Figure 1. The findings reveal that the general trends observed in 7B models persist in these larger models.
    }
    \label{fig:llama_qwen}
\end{figure}

\subsection{Comparison Across Different LLMs (RQ2)}

In this section, we analyze how various LLMs perform across multiple languages using the Language Ranker. We focus on understanding the consistency of language performance among different LLMs, including models with varying architectures and training specifics. We have the following findings:

\label{3.2}
\noindent
(1) \textbf{Different models display similar results across languages}.   
Figure~\ref{fig:main_label} presents the cosine similarity scores across various layers for four different 7B parameter LLMs: LlaMa2, Gemma, Mistral, and Qwen. Four LLMs display a similar trend where high-resource languages (i.e., German, Spanish, French, Malay, and Chinese) consistently exhibit higher cosine similarity scores compared to low-resource languages (i.e., Igbo, Kazakh, Kannada, Oriya, and Turkmen).
Figure \ref{fig:heat} further corroborates these findings by comparing the rank correlation of the similarity scores across the LLMs. Each LLM's ranking is used as a baseline, and the remaining three models exhibit ranking patterns that are largely similar to this baseline. This similarity indicates that despite differences in model architecture or training specifics, the relative performance of languages remains consistent across these four models.

\noindent
(2) \textbf{Fine-tuning on specific languages will improve its performance}. According to the technical report of Qwen~\cite{bai2023qwen}, Qwen has additional fine-tuning on the Chinese corpus, which leads to better performance in Chinese. 
In Figure 1, we observe that for LlaMa2, Gemma, and Mistral, the performance of Chinese is slightly lower than that of other high-resource languages. However, for Qwen, the performance of Chinese is roughly comparable to other high-resource languages and even shows a gradual improvement in the last few layers. This improvement is more clear in Qwen, mainly due to additional fine-tuning of the Qwen model family on the Chinese corpus, as noted in the technical report of Qwen.

\noindent
(3) \textbf{Comparison of Larger LLM Sizes.}
We extended our analysis to explore the performance of the similarity metric in larger language models with 13 billion parameters. Figure~\ref{fig:llama_qwen} presents the results for LlaMa2 13B and Qwen 13B. The findings reveal that the general trends observed in 7B models persist in these larger models. Specifically, the relative performance of high-resource and low-resource languages remains largely unchanged, with a clear separation between the two groups. LlaMa2 13B exhibits more pronounced fluctuations in the initial layers compared to its 7B counterpart, suggesting potentially richer early-stage language representations. 
These observations highlight both the consistency of language performance patterns across model sizes and the potential for larger models to enhance representations for specific languages.

\begin{table}[t]
\centering
\resizebox{\columnwidth}{!}{
\begin{tabular}{lrr|lrr}
\hline
Language & \multicolumn{1}{l}{Proportion}&\multicolumn{1}{l}{Similarity}& Language & \multicolumn{1}{l}{Proportion}&\multicolumn{1}{l}{Similarity} \\
\hline  
German & 0.17\%  &0.723 &  Welsh&$\leq$0.01\% &0.396\\
French & 0.16\%  &0.737& Persian &$\leq$0.01\% &0.300\\
Swedish& 0.15\% &0.662& Urdu&$\leq$0.01\%   &0.275\\
Chinese&0.13\%  &0.552&  Kannada &$\leq$0.01\% &0.236\\
\hline
\end{tabular}
}
\caption{The proportion of different languages in the LlaMa2 pre-training corpus and the similarity metric we proposed. The English language ratio is 89.7\%.}
\label{tab:lid}
\end{table}

\begin{table*}
\centering
\scalebox{0.85}{
\begin{tabular}{lcccccccc}
\toprule
\textbf{Language}& \multicolumn{4}{c}{\textbf{ARC}} & \multicolumn{4}{c}{\textbf{MMLU}}\\
\cmidrule(lr){2-5}\cmidrule(lr){6-9} & LlaMa2 7B & Gemma 7B & Mistral 7B &  Qwen 7B& LlaMa2 7B & Gemma 7B & Mistral 7B &  Qwen 7B\\
\hline
Chinese&27\% &71\%&57\%&66\% & 32\%&54\% & 37\%& 44\%\\
German&27\% &68\% &63\% &32\% & 25\%& 57\%& 47\%& 27\%\\
French&31\% &76\% &59\% &42\% & 24\%& 58\%& 48\%& 28\%\\
Spanish&31\% &77\% &60\% &46\% & 29\%& 56\%& 52\%& 33\%\\
Italian&29\% &77\% &67\% &44\% & 23\%& 56\%& 44\%&32\% \\
\hline 
Kannada&24\% &48\% &27\% &21\% & 21\%& 40\%& 22\%& 19\%\\
Hindi&28\% &60\% &42\% &22\% &25\% & 45\%& 32\%& 23\%\\
Armenian&19\% &40\% &36\% &20\% & 20\%& 36\%& 30\%& 25\%\\
Marathi&28\% &46\% &26\% &25\% & 27\%& 42\%&30\% &26\% \\
Telugu&30\% &42\% &30\% &30\% & 24\%& 33\%&34\% & 23\%\\
\bottomrule
\end{tabular}}
\caption{Performance on two inference tasks}
\label{tab:inference_table}
\end{table*}

\begin{figure*}[t]
    \centering
    \vspace{-10pt}
    \includegraphics[width=1.0\linewidth]{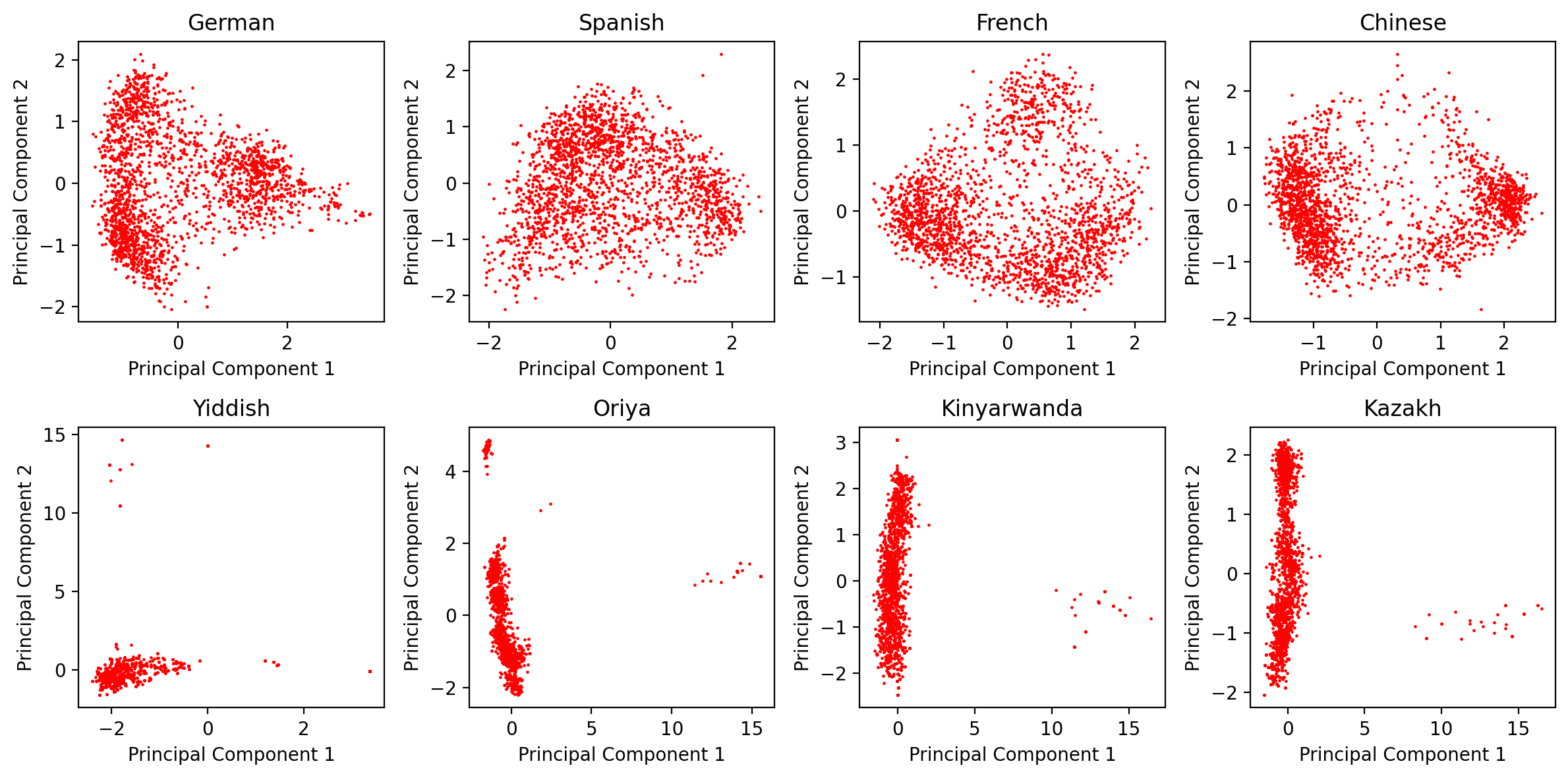}
    \caption{Visualization of the embedding space of Gemma 7B for eight languages. Four figures at the top are high-resource languages, Four figures at the bottom are low-resource languages. }
    \label{fig:single_cluster}
\end{figure*}

\subsection{Relationship to Ratio of Training Corpus? (RQ3)}
\label{3.3}

In this section, we explore the relationship between the proportion of each language in the LlaMa2 pre-training corpus and their corresponding performance as measured by the similarity metric.
According to the technical report of LlaMa2~\cite{touvron2023llama}, we obtain the proportion of the pre-training corpus of some languages. 
Table~\ref{tab:lid} illustrates this relationship by listing a selection of languages with their proportion in the training data and their similarity scores relative to English. The table is divided into two parts: the left side lists high-resource languages with relatively higher proportions in the LlaMa2 pre-training corpus, and the right side lists low-resource languages with very low proportions ($\leq$ 0.01\%). For example, German, with a proportion of 0.17\%, has a high similarity score of 0.723, indicating strong performance in comparison to English.  
This trend suggests that languages with a higher proportion in the pre-training corpus tend to have higher similarity scores, reflecting better model performance. In contrast, low-resource languages like Kannada and Urdu, each with proportions of less than 0.01\%, have much lower similarity scores (i.e., 0.236 and 0.275).

\begin{table*}[]
\centering
\resizebox{2\columnwidth}{!}{
\begin{tabular}{ll|ll|ll}
\hline
High-High & \multicolumn{1}{l}{Similarity Score}& High-Low & \multicolumn{1}{l}{Similarity Score} & Low-Low & \multicolumn{1}{l}{Similarity Score}\\
\hline
English-German & 0.72 &German-Silesian &0.48&Azerbaijani-Turkmen&0.51\\
Italian-French & 0.68 &French-Erzya&0.35&Hungarian-Yiddish&0.24\\
German-French & 0.67&Italian-Romany&0.32&Kab-SMT&0.36\\
French-Chinese&0.59&Italian-Uighur&0.27&Mari-Tatar&0.48\\
\hline
\end{tabular}
}
\caption{Similarity score of different language pairs of Gemma 7B.}
\label{tabel:gemma H-L}
\end{table*}

\begin{table*}[]
\centering
\resizebox{2\columnwidth}{!}{
\begin{tabular}{ll|ll|ll}
\hline
High-High & \multicolumn{1}{l}{Similarity Score}& High-Low & \multicolumn{1}{l}{Similarity Score} & Low-Low & \multicolumn{1}{l}{Similarity Score}\\
\hline
English-German & 0.72 &German-Silesian &0.44&Azerbaijani-Turkmen&0.51\\
Italian-French & 0.69 &French-Erzya&0.31&Hungarian-Yiddish&0.19\\
German-French & 0.68&Italian-Romany&0.15&Kab-SMT&0.40\\
French-Chinese&0.56&Italian-Uighur&0.20&Mari-Tatar&0.42\\
\hline
\end{tabular}
}
\caption{Similarity score of different language pairs of LlaMa2 7B.}
\label{tabel:llama H-L}
\end{table*}

\subsection{Correlation with Other Inference Tasks? (RQ4)}
\label{3.4}
To more comprehensively reflect the performance of LLMs in various languages, we evaluate the multilingual reasoning ability of these LLMs.

We use MLMM-evaluation\footnote{\scriptsize{\url{https://github.com/nlp-uoregon/mlmm-evaluation}}} 
as our benchmark dataset to evaluate LLMs' performances on reasoning tasks in various languages. The benchmark dataset can be used to evaluate the LLM across 26 different languages. It consists of three datasets: ARC~\cite{clark2018think}, HellaSwag~\cite{zellers2019hellaswag}, and MMLU~\cite{hendrycks2020measuring}. We chose ARC and MMLU for evaluation, and both of them are multiple-choice datasets. The ARC dataset consists of 7,787  multiple-choice science questions drawn from a variety of sources. The MMLU dataset contains multiple-choice questions derived from diverse fields of knowledge. 
We selected five high-resource languages (Chinese, German, French, Spanish, Italian) and five low-resource languages (Kannada, Hindi, Armenian, Marathi, Telugu) for evaluation, randomly selected 100 samples from each language for 4-shot learning prediction, and used accuracy as the metric.
The LLMs evaluated are consistent with Figure~\ref{fig:main_label}: LlaMa2 7B, Gemma 7B, Mistral 7B, and Qwen 7B.

The predicted result is shown in Table~\ref{tab:inference_table}. From the result, we find that for Gemma, Mistral, and Qwen, the performance of high-resource languages is significantly better than that of low-resource languages, and Gemma performs best. For the LlaMa2, the performance in all languages is generally not as good as the first three LLMs. This result shows that LLM reasoning ability in low-resource languages is worse than that in high-resource languages.
This result proves that there are differences in performance between high-resource and low-resource languages in reasoning tasks, illustrating the effectiveness of the proposed cosine similarity metric.

\section{Further Analysis of Proposed Metric} 
In the last section, we introduced and evaluated the Language Ranker, demonstrating its ability to quantify the multilingual capabilities of LLMs by comparing their internal representations against an English baseline. This provided a robust measure of how LLMs perform across different languages, especially highlighting the disparities between high-resource and low-resource languages.

Building on these insights, in this section, we delve deeper into the proposed metric to explore its credibility and reliability further. Specifically, we aim to answer the following questions in the following three subsections: RQ5: Is choosing English as the benchmark a wise choice?
RQ6: What does the subspace of each language look like?
RQ7: Is choosing cosine similarity a wise choice?

\begin{figure*}
    \centering
    
    \includegraphics[width=0.9\linewidth, trim=0 0 0 0]{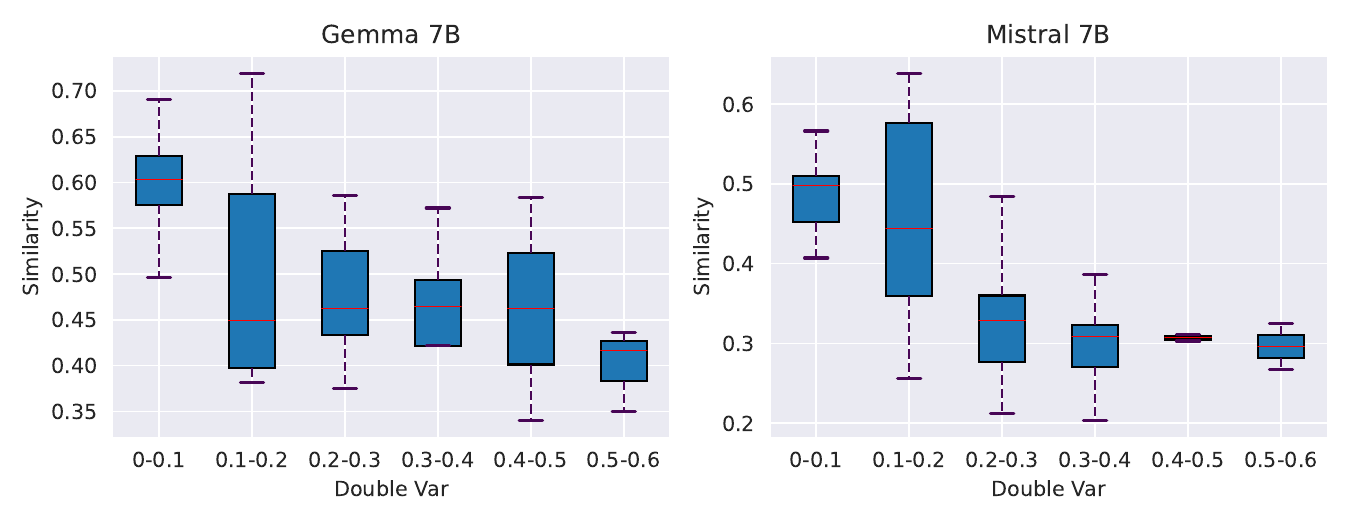}
    \caption{Box plot of the relationship between the double variance and similarity of Gemma 7B and Mistral 7B.}
    \label{fig:double_var}
\end{figure*}

\subsection{Why Using English as Baseline? (RQ5)}\label{eng}
In the above sections, we choose English as a baseline. This is based on the a priori assumption that low-resource languages generally perform worse than high-resource languages. But if we choose other high-resource languages as baselines, will we get the same performance? In other words, how can we ensure that our metric is not affected by the English language itself? To answer this question, we divided our probing datasets into three types: High Resource-High Resource (\textbf{H-H}), High Resource-Low Resource (\textbf{H-L}), and Low Resource-Low Resource (\textbf{L-L}). To fulfill our requirement, we utilize Tatoeba-Challenge~\cite{tiedemann-2020-tatoeba} as our dataset instead of opus-100 because the latter is an English-centric dataset which means there is no Low Resource-Low Resource language pair. Tatoeba-Challenge is a challenge set for machine translation that contains 32G translation units in 2,539 bitexts. The whole data set covers 487 languages linked to each other in 4,024 language pairs. We select four language pairs for each group, English-German (en-de), English-French (en-fr), German-French (de-fr), and Italian-German (it-de) represent \textbf{H-H}; German - Silesian (de-szl), French-Erzya (fr-myv), Italian - Romany (it-ro) and Italian - Uighur (it-uig) represent \textbf{H-L}; Azerbaijani-Turkmen (az-tr), Hungarian-Yiddish (hu-yi), Kabyle-Standard Moroccan Tamazight (kab-SMT) and Mari-Tatar (ma-ta) represent \textbf{L-L}.
The results are shown in Table \ref{tabel:gemma H-L} and Table \ref{tabel:llama H-L}.

From the results, we can observe that the score of High-High is higher than the score of High-Low and Low-Low universally. An obvious inference is that the distribution of high-resource languages is relatively close to each other, while the distribution of low-resource languages varies greatly, neither being close to each other nor to high-resource languages. 
Therefore, the distribution of high-resource languages is relatively consistent, while the distribution of low-resource languages varies greatly. Choosing English as the baseline is a convenient choice. Thus, we can also choose other high-resource languages such as German and French as the baseline.

\subsection{Deeper Analysis of the Embedding Space (RQ6)}\label{embedd}
We explained why we chose English as the baseline in the above section. Choosing English is actually choosing a high-resource language as our baseline. A question naturally arises: How can we make sure that the performance of high-resource languages is better than the performance of low-resource languages? The result of Table~\ref{tab:inference_table} has confirmed the answer to the question by the reasoning task. To answer this question more deeply, we need to analyze the distribution of the embedding of different languages. As is shown in Figure \ref{fig:single_cluster}, the top four sub-figures are embedding spaces of high-resource languages, while the bottom four sub-figures are embedding spaces of low-resource languages. It is obvious that high-resource languages are more evenly distributed throughout the space, while low-resource languages are more narrowly distributed, and compressed into a near straight line. Therefore, the performance of low-resource languages is worse than that of high-resource languages, which means it is suitable to choose a high-resource language like English as the baseline.

\begin{table*}[htbp]
\centering
\resizebox{\textwidth}{!}{ 
\begin{tabular}{lrr|lrr|lrr|lrr}
\hline
\multicolumn{6}{c|}{\textbf{Gemma 7B}} & \multicolumn{6}{c}{\textbf{Mistral 7B}} \\
\hline
\multicolumn{3}{c|}{\textbf{High-resource}} & \multicolumn{3}{c|}{\textbf{Low-resource}} & 
\multicolumn{3}{c|}{\textbf{High-resource}} & \multicolumn{3}{c}{\textbf{Low-resource}} \\
\hline
\textbf{Language} & \textbf{Double Var} & \textbf{Similarity} & 
\textbf{Language} & \textbf{Double Var} & \textbf{Similarity} &
\textbf{Language} & \textbf{Double Var} & \textbf{Similarity} & 
\textbf{Language} & \textbf{Double Var} & \textbf{Similarity} \\
\hline
Italian & 0.04 & 0.66 & Nepali & 0.75 & 0.42 & Italian & 0.12 & 0.59 & Nepali & 0.60 & 0.28 \\
French & 0.09 & 0.69 & Kazakh & 0.85 & 0.38 & French & 0.17 & 0.65 & Kazakh & 0.32 & 0.32 \\
Spanish & 0.06 & 0.68 & Burmese & 0.36 & 0.30 & Spanish & 0.17 & 0.64 & Burmese & 0.40 & 0.20 \\
German & 0.10 & 0.72 & Pashto & 0.72 & 0.36 & German & 0.19 & 0.66 & Pashto & 0.73 & 0.29 \\
\hline
\end{tabular}
}
\caption{Similarity scores and double variance results for some languages on Gemma 7B and Mistral 7B.}
\label{tab:combined_double}
\end{table*}

\subsection{Why Using Cosine Similarity? (RQ7)}\label{cosine}
Recent research~\cite{Steck_2024} has shown that cosine similarity is not always a reliable metric. 
Inspired by Section~\ref{embedd}, the quality of the performance of various languages can be clearly judged from the subspace distribution. Therefore, we decided to quantify the performance of these languages from a distribution perspective. 

Back to Figure \ref{fig:single_cluster}, the projection distributions of high-resource languages in different directions are relatively consistent, while the distribution of low-resource languages is compressed approximately into a straight line. The projection distributions outside the straight line, such as the projections perpendicular to the straight line, are crowded in a smaller area. This suggests that we can use the projection variance to approximately measure the quality of the distribution.

According to PCA, we assume the embedding vectors are $\{X_i\}_{i=1}^{n}$(being centralized), the projection direction is $\omega$, the project variance $Var(X,\omega)$ can be calculated as follows:
\begin{equation}
\small
\begin{aligned}
    Var(X,\omega) &= \frac{1}{n} \sum_{i=1}^{n} (X_{i}^T\omega)^2 = \frac{1}{n} \sum_{i=1}^{n}  \omega^T X_i X_{i}^T \omega\\
                  & = \omega^T Cov(X) \omega
    \quad s.t. \quad \omega^T \omega = 1
\end{aligned}
\end{equation}
It is obvious that $Var(X,\omega)$ is the eigenvalue of the $Cov(X)$. For high-resource languages, projection variance in different directions should be close to each other so that it can be evenly distributed in all directions. The opposite is true for low-resource languages. Therefore, we can extract the first K eigenvalues $\{Var(X,\omega_i)\}_{i=1}^{k}$ and calculate their variance. The variance of the eigenvalues can be used to measure the differences in the distribution in each direction, which is called double variance.  This metric can be used to specifically measure the quality of the distribution. The higher the double variance, the more unbalanced the distribution and the worse the performance, and the vice versa.

We employ the box plot to show the relationship between the proposed cosine similarity metric and the double variance metric more clearly.
From Table~\ref{tab:combined_double}, we can observe that for each LLM, the languages in the left part are some common high-resource languages, which have higher similarity and lower double variance, while the right part is the opposite for low-resource languages. 
The second observation is that as the variance increases, the similarity score also tends to decrease. 
The increase in variance means that the distribution of the subspace becomes uneven and the similarity score decreases accordingly. This shows that the proposed cosine similarity metric can be utilized to roughly measure the quality of distribution of the subspace, which can thus measure the performance of LLM in different languages.

\section{Related Work}

\noindent\textbf{Representation Engineering.}\,
Representation engineering has emerged as an effective approach to enhance the interpretability and transparency of LLMs. Researchers have been leveraging internal representations to tackle various challenges. \citet{zou2023representation} summarizes the application of representation engineering in bias, fairness, model editing, and other areas. 
\citet{gurnee2023language} found that the internal representation of LLM has a certain correlation with time and space, and the internal representation can be employed to represent time and space. 
\citet{li2024inference} found that the representation of the attention head inside the LLM can be used to indicate the correct reasoning direction, probe analysis is further used to correct the internal representation direction to improve the LLM's performance.
\citet{marks2023geometry} study the structure of LLM representations of true/false statements,  proved that language models linearly represent the true/false of factual statements. \citet{ju2024large} used probe technique to detect how LLM stores knowledge layer by layer.

\noindent
\textbf{Multilingual Language Model.}\, 
Recent research such as \cite{qin2024multilingual} summarizes the recent progress and future trends in multilingual large language models.
\citet{ahuja2024megaverse} constructed a benchmark to evaluate LLM's multilingual ability comprising 22 datasets covering 83 languages.
\citet{huang2023not,qin2023cross} have proven that LLM performance varies substantially across different languages and they employ a prompt technique to improve task performance across languages. \citet{wendler2024llamas} explores how LlaMa2 works in multilingual tasks and what role English plays in these tasks.
The imbalance distribution of training corpus in different languages leads to the bias of LLM towards some high-resource languages such as English~\cite{blasi2021systematic}. Some approaches employ multilingual language modeling to alleviate the phenomenon \cite{shen2024language,kalyan2021ammus,conneau2019unsupervised}. These studies show the importance of strengthening the cross-lingual capabilities of the pre-trained model. \citet{schafer2024language} found that the presence of a primary language in the training process of LLMs can improve the performance of low-resource languages and lead to a more consistent representation of LLMs in different languages. \citet{liu2024translation}  found that for English-centric LLMs, although translation into English helps improve the performance of NLP tasks, it is not the best choice for all situations.

\section{Conclusions and Future Work}

In this work, we propose the Language Ranker to evaluate the performance of LLMs across diverse languages by comparing their internal representations to English.
The results show that high-resource languages show higher similarity scores with English, while low-resource languages have lower scores, validating the effectiveness of our metric in assessing language performance. Besides, there is a strong correlation between the performance of LLMs in different languages and the proportion of those languages in the pre-training corpus. 
Further, results indicate that high-resource languages are more evenly distributed in the embedding space, whereas low-resource languages tend to be narrowly clustered. 
In the future, we plan to design more comprehensive benchmarks to measure LLM's capabilities in different languages. Besides, we plan to explore the application of the Language Ranker to guide the development of more balanced multilingual training datasets and improve LLM performance on low-resource languages.

\clearpage
\bibliography{aaai25,anthology}

\clearpage
\appendix
\section{Appendix}
\label{sec:appendix}

\subsection{Ranking Result For LLMs}
 We give the similarity scores of the four LLMs used in the experiment on 18 high-resource languages. Results are shown in the following tables.
\begin{table}[!h]
\centering
\resizebox{\columnwidth}{!}{
\begin{tabular}{ll|ll}
\hline
Language & \multicolumn{1}{l}{Similarity Score}& Language & \multicolumn{1}{l}{Similarity Score} \\
\hline
    
German & 0.723 &  Western Frisian&0.378\\
French & 0.737 & Tamil & 0.347 \\
Spanish & 0.768  &Gujarati & 0.313 \\ 
Italian & 0.706  & Kurdish & 0.308 \\
Russian & 0.734 &Pashto & 0.284 \\
Dutch & 0.709 & Assamese & 0.260 \\
Polish & 0.664&Central Khmer & 0.240 \\
Malay & 0.651 & Panjabi & 0.218 \\
Swedish & 0.661 &Amharic & 0.202 \\
\hline
\end{tabular}
}
\label{tabel:llama socre}
\caption{The similarity score of LlaMa2 7B.}
\end{table}

\begin{table}[!h]
\centering
\resizebox{\columnwidth}{!}{
\begin{tabular}{ll|ll}
\hline
Language & \multicolumn{1}{l}{Similarity Score}& Language & \multicolumn{1}{l}{Similarity Score} \\
\hline
German & 0.719 &  Western Frisian&0.443\\
French & 0.691 & Tamil & 0.420 \\
Spanish & 0.683  &Gujarati & 0.433 \\ 
Italian & 0.662  & Kurdish & 0.358 \\
Russian & 0.674 &Pashto & 0.362 \\
Dutch & 0.658 & Assamese & 0.396 \\
Polish & 0.618&Central Khmer & 0.330 \\
Malay & 0.615 & Panjabi & 0.379 \\ 
Swedish & 0.629 &Amharic & 0.298 \\
\hline
\end{tabular}
}
\label{tabel:gemma socre}
\caption{The similarity score of Gemma 7B.}
\end{table}

\begin{table}[htbp]
\centering
\resizebox{\columnwidth}{!}{
\begin{tabular}{ll|ll}
\hline
Language & \multicolumn{1}{l}{Similarity Score}& Language & \multicolumn{1}{l}{Similarity Score} \\
\hline
German & 0.639 &  Western Frisian&0.346\\
French & 0.623 & Tamil & 0.279 \\
Spanish & 0.616  &Gujarati & 0.270 \\ 
Italian & 0.571  & Kurdish & 0.262 \\
Russian & 0.611 &Pashto & 0.267 \\
Dutch & 0.566 & Assamese & 0.276 \\
Polish & 0.514&Central Khmer & 0.252 \\
Malay & 0.497 & Panjabi & 0.213 \\ 
Swedish & 0.532 &Amharic & 0.191 \\
\hline
\end{tabular}
}
\label{tabel:mistral socre}
\caption{The similarity score of Mistral 7B.}
\end{table}

\begin{table}[!h]
\centering
\resizebox{\columnwidth}{!}{
\begin{tabular}{ll|ll}
\hline
Language & \multicolumn{1}{l}{Similarity Score}& Language & \multicolumn{1}{l}{Similarity Score} \\
\hline
German & 0.805 &  Western Frisian&0.441\\
French & 0.793 & Tamil & 0.510 \\
Spanish & 0.800  &Gujarati & 0.469 \\ 
Italian & 0.773  & Kurdish & 0.436 \\
Russian & 0.794 &Pashto & 0.448 \\
Dutch & 0.773 & Assamese & 0.507 \\
Polish & 0.752&Central Khmer & 0.407 \\
Malay & 0.730 & Panjabi & 0.385 \\ 
Swedish & 0.759 &Amharic & 0.470 \\
\hline
\end{tabular}
}
\label{tabel:qwen socre}
\caption{The similarity score of Qwen 7B.}
\end{table}

\begin{figure*}[t]
    \centering
    \vspace{-10pt}
    \includegraphics[width=1.0\linewidth]{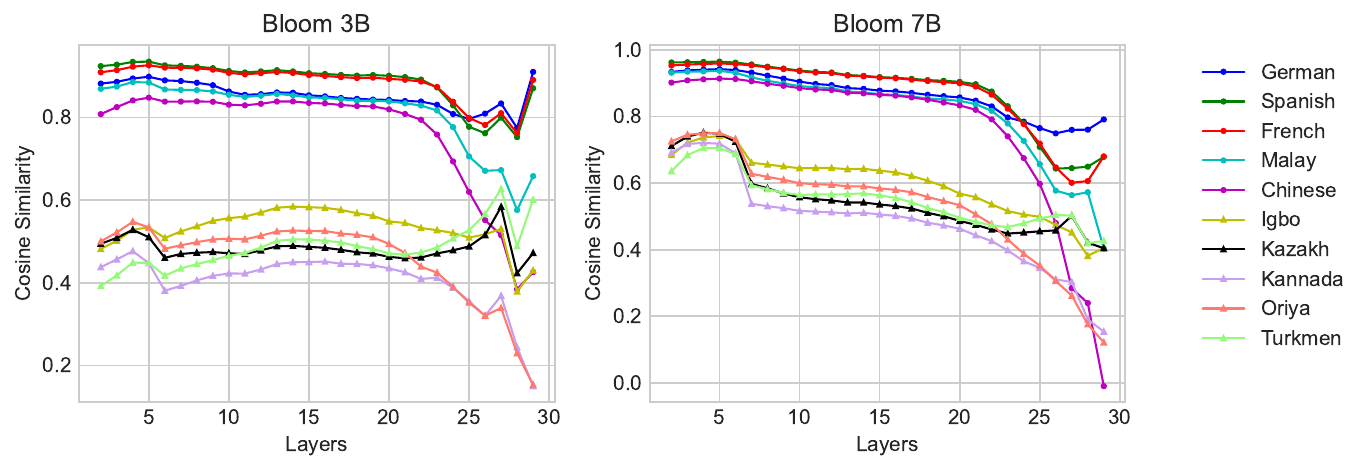}
    \caption{Simlarity scores curves of Bloom 3B and Bloom 7B. }
    \label{fig:bloom}
\end{figure*}

\subsection{Details of experiment in Table~\ref{tabel:llama H-L} and Table~\ref{tabel:gemma H-L}}
We selected data from the Tatoeba-Challenge repository\footnote{https://github.com/Helsinki-NLP/Tatoeba-Challenge/tree/master/data}. Since the number of samples for some low-resource language pairs is small, we extracted 100 samples for each language pair. If there are less than 100, we extracted all samples and extracted them according to the test-dev-train priority.
We can observe that the scores of language pairs of the same type may vary differently but the performance of each LLM on different types of language pairs is roughly the same. 
\begin{table}[htbp]
\centering
\resizebox{1\columnwidth}{!}{
\begin{tabular}{ll|ll|ll}
\hline
High-High & \multicolumn{1}{l}{Similarity Score}& High-Low & \multicolumn{1}{l}{Similarity Score} & Low-Low & \multicolumn{1}{l}{Similarity Score}\\
\hline
English-German & 0.64 &German-Silesian &0.43&Azerbaijani-Turkmen&0.51\\
Italian-French & 0.60 &French-Erzya&0.30&Hungarian-Yiddish&0.18\\
German-French & 0.62&Italian-Romany&0.25&Kab-SMT&0.39\\
French-Chinese&0.57&Italian-Uighur&0.24&Mari-Tatar&0.46\\
\hline
\end{tabular}
}
\caption{Similarity score of different language pairs of mistral 7B.}
\label{tabel:mistral H-L}
\end{table}

\begin{table}[htbp]
\centering
\resizebox{1\columnwidth}{!}{
\begin{tabular}{ll|ll|ll}
\hline
High-High & \multicolumn{1}{l}{Similarity Score}& High-Low & \multicolumn{1}{l}{Similarity Score} & Low-Low & \multicolumn{1}{l}{Similarity Score}\\
\hline
English-German & 0.81 &German-Silesian &0.58&Azerbaijani-Turkmen&0.67\\
Italian-French & 0.75 &French-Erzya&0.51&Hungarian-Yiddish&0.39\\
German-French & 0.80&Italian-Romany&0.53&Kab-SMT&0.55\\
French-Chinese&0.71&Italian-Uighur&0.49&Mari-Tatar&0.55\\
\hline
\end{tabular}
}
\caption{Similarity score of different language pairs of qwen 7B.}
\label{tabel:qwen H-L}
\end{table}

\subsection{Performance of LLMs with Other Sizes}

\subsubsection{Bloom 3B and Bloom 7B}
Figure~\ref{fig:bloom} shows the result of Bloom 3B and Bloom 7B. Except for the last few layers of the model, there are obvious differences between high-resource languages and low-resource languages which are similar to the above LLMs, while there are smaller differences within the same category of languages. The score is higher than LlaMa2, Gemma, Mistral, and Qwen.

\subsection{The Relation Between the Performance of Reasoning  and The Similarity Score}

\begin{figure}[htbp]  
  \centering  
  \includegraphics[width=0.5\textwidth]{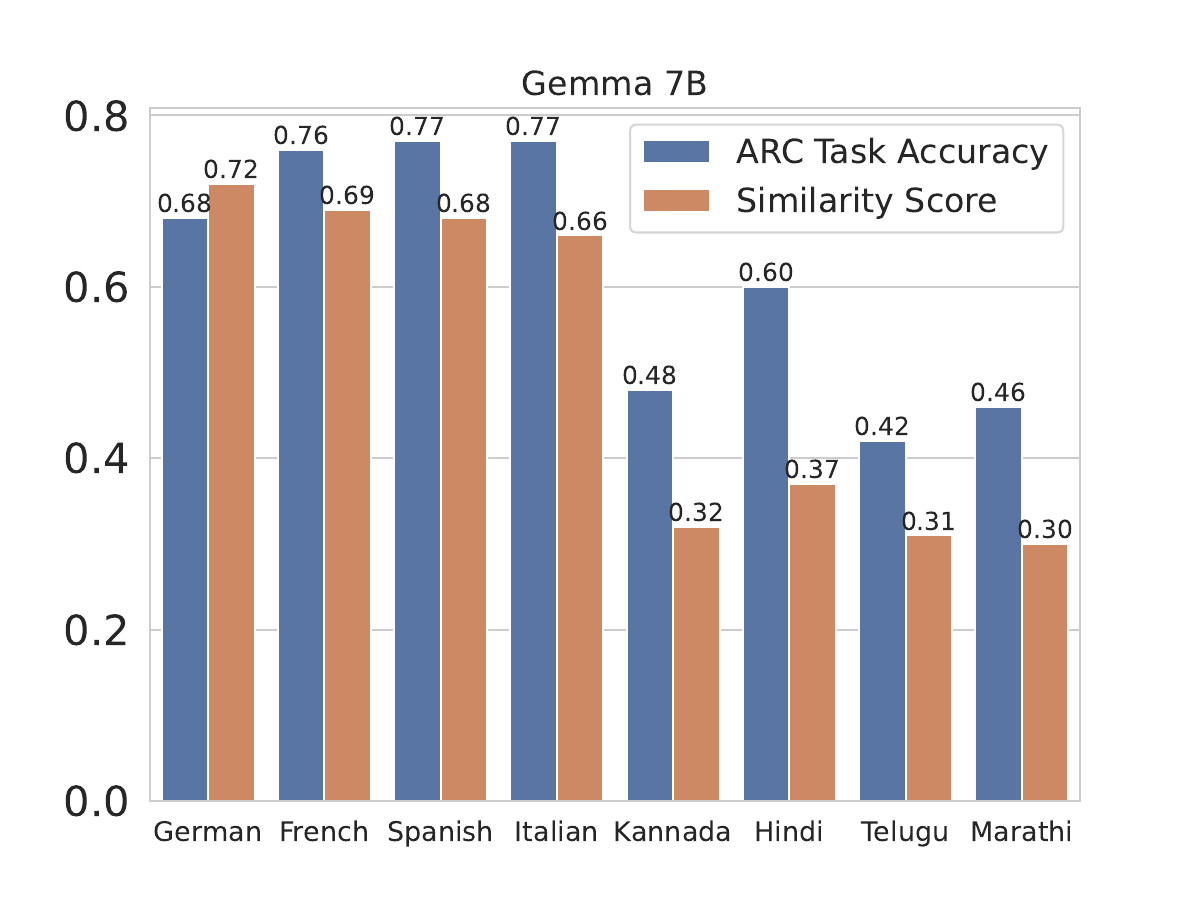}  
  \caption{Accuracy of ARC reasoning tasks and similarity scores  in different languages on Gemma 7B.}  
  \label{fig:gemma_rel}  
\end{figure}

\begin{figure}[htbp]  
  \centering  
  \includegraphics[width=0.5\textwidth]{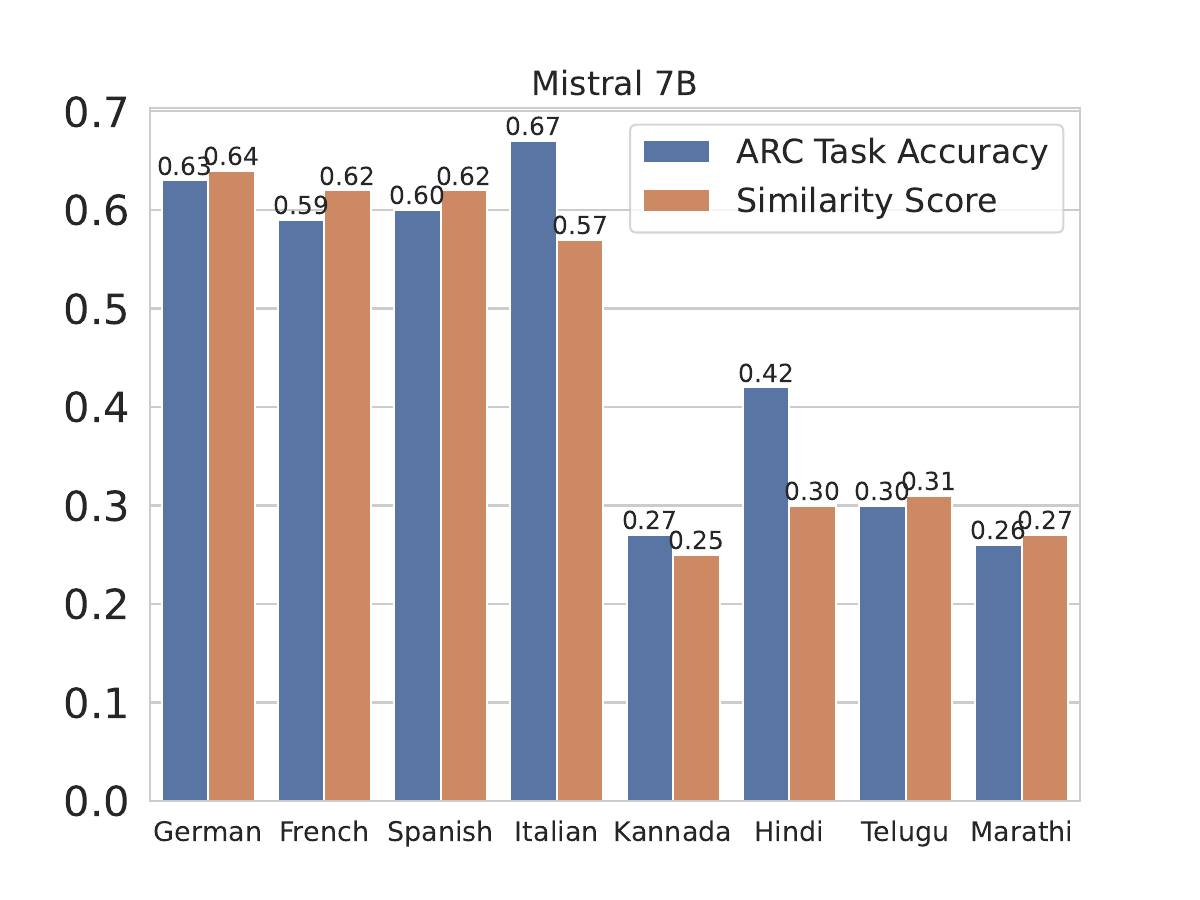}  
  \caption{Accuracy of ARC reasoning tasks and similarity scores  in different languages on Mistral 7B.} 
  \label{fig:mistral_rel}  
\end{figure}

We also quantify the relationship between the performance of the inference task (Table~\ref{tab:inference_table}) and the language similarity to English (Figure~\ref{fig:main_label}).
It can be clearly observed from Figure~\ref{fig:gemma_rel} and Figure~\ref{fig:mistral_rel} that the performance of LLM in reasoning tasks is highly correlated with the similarity score, it performs better in high-resource languages(the first four languages) and worse in low-resource languages(the last four languages).

\newpage

\end{document}